\documentclass[10pt,letterpaper]{amsart}
\usepackage[margin=1.25in]{geometry}
\usepackage[foot]{amsaddr}
\usepackage{times}
\usepackage{epsfig}
\usepackage{graphicx}
\usepackage{subcaption}
\usepackage{amsmath}
\usepackage{algorithm}
\usepackage{algorithmic}
\usepackage{amssymb}
\usepackage{adjustbox}
\usepackage{dblfloatfix}
\usepackage{multirow}

\newcommand{\R}{\mathbb{R}}
\newcommand{\cX}{\mathcal{X}}

\newcommand\B{\rule[-1.2ex]{0pt}{0pt}}
\DeclareMathOperator*{\argmax}{arg\,max}

\usepackage{siunitx}
\graphicspath{{Figs/}}
\begin{document}

%%%%%%%%% TITLE
\title[The LogBarrier adversarial attack]{The LogBarrier adversarial attack: making effective use of decision boundary
  information
}
\author{Chris Finlay$^1$ \and Aram-Alexandre Pooladian$^1$ \and Adam M. Oberman$^{1,2}$}
\address{$^1$ Department of mathematics and statistics, McGill University}
\address{$^2$ AIFaster Consulting Inc}
\thanks{AO supported by AFOSR grant FA9550-18-1-0167}
\email{christopher.finlay@gmail.com}
\email{aram-alexandre.pooladian@mail.mcgill.ca}
\email{adam.oberman@mcgill.ca}
%\email{https://www.adamoberman.net/}
%\urladdr{https://www.adamoberman.net/}
\date{\today}

%\author{
%Chris Finlay$^*$\\
%McGill University\\
%{\tt\small christopher.finlay@mail.mcgill.ca}
%% omit the following lines up until the closing ``}''.
%% Additional authors and addresses can be added with ``\and'',
%% just like the second author.
%% To save space, use either the email address or home page, not both
%\and
%Aram-Alexandre Pooladian$^*$\\
%McGill University\\
%{\tt\small aram-alexandre.pooladian@mail.mcgill.ca}
%\and 
%Adam Oberman\\
%McGill University\\
%{\tt\small adam.oberman@mcgill.ca}
%}
%

\maketitle
%\thispagestyle{empty}
%%%%%%%%% ABSTRACT
\begin{abstract}
Adversarial attacks for image classification are small perturbations to images
that are designed to cause misclassification by a model.  Adversarial
attacks formally correspond to an optimization problem: find a minimum norm
image perturbation, constrained to cause misclassification.   A number of
effective attacks have been developed.  However, to date, no gradient-based attacks have
used best practices from the optimization literature to solve this constrained  minimization
problem. We design a new
untargeted attack, based on these best practices, using the well-regarded logarithmic barrier
method.

On average, our attack distance is similar or better than all state-of-the-art
attacks on benchmark datasets  (MNIST, CIFAR10, ImageNet-1K).  In addition, our
method performs significantly better on the most challenging images, those which
normally require larger perturbations for misclassification.  We employ the
LogBarrier attack  on several adversarially defended models, and show that it
adversarially perturbs all images more efficiently than other attacks: the
distance needed to perturb all images is significantly smaller with the
LogBarrier attack than with other state-of-the-art attacks.
\end{abstract}

%%%%%%%%% BODY TEXT
\section{Introduction}
Deep learning models have achieved impressive results in many areas of
application. However, deep learning models
remain vulnerable to adversarial attacks \cite{szegedy2013}: small changes (imperceptible to the human eye)
in the model input may
lead to vastly different model predictions.  In security-based applications,
this vulnerability is of utmost concern. For
example, traffic signs may be modified with
small stickers to cause misclassification, causing say a stop sign to be
treated as speed limit sign \cite{eykholt2018}.  Facial recognition systems can be easily spoofed
using colourful glasses \cite{sharif2016}.

This security flaw has led to an arms race in the research community, between
those who develop defences to
adversarial attacks, and those working to overcome these defences with stronger
adversarial attack
methods \cite{convexpolytope,raghunathan2018}. Notably, as the community develops
stronger adversarial attack methods, claims of model robustness to adversarial
attack are often proved to be
premature \cite{carlini2017,athalye2018}. 

There are two approaches to demonstrating a model is resistant to adversarial
attacks. The first is \emph{theoretical}, via a provable lower bound on the minimum
adversarial distance necessary to cause misclassification
\cite{convexpolytope,raghunathan2018,katz2017}. Theoretical lower bounds are often
pessimistic: the gap between the theoretical lower bound and adversarial
examples generated by state-of-the-art attack algorithms can be large. 
Therefore, a second \emph{empirical} approach is also used: an upper bound on the
minimum adversarial distance is demonstrated through adversarial examples
created by adversarial attacks \cite{kurakin2016,madry2017,carlini2017,boundaryattack}.  
The motivation to design \emph{strong} adversarial attacks is therefore twofold: on the
one hand, to validate theoretical lower bounds on robustness; and on the other,
to construct empirical upper bounds on the minimum adversarial distance.
Ideally, the gap between the theoretical lower bound and the empirical upper
bound should be small. As adversarial attacks become stronger, the gap narrows
from above.

The process of finding an adversarial example with an adversarial attack is
an optimization problem: find a small perturbation of the model input which
causes misclassification. This optimization problem has been recast in
various ways.  Rather than directly enforcing misclassification, many
adversarial attacks instead attempt to maximize the loss function. The Fast
Gradient Signed Method (FGSM) was one of the first adversarial attacks to do so
\cite{szegedy2013}, measuring perturbation size in the $\ell_\infty$ norm.
Iterative versions of FGSM were soon developed, where perturbations were
measured in either the
$\ell_\infty$ and $\ell_2$ norms \cite{kurakin2016, madry2017, zheng2018}. These
iterative methods perform Projected Gradient Descent (PGD), maximizing the loss
function subject to a constraint enforcing small perturbation in the appropriate
norm. Other works have studied sparse adversarial attacks, as in
\cite{papernot2016}. Rather than maximizing the loss, Carlini and Wagner
\cite{carlini2017} developed a strong adversarial attack by forcing
misclassification to a predetermined target class. If only the decision of the
model is available (but not the loss or the model gradients), adversarial
examples can still be found using gradient-free optimization techniques
\cite{boundaryattack}.

In this paper, rather than using the loss as a proxy for misclassification, we
design an adversarial attack that solves the adversarial optimization
problem \emph{directly}: minimize the size of the input perturbation subject to a
misclassification constraint. Our method is gradient-based, but does not use
the training loss function. The method is based on a sound, well-developed
optimization technique, namely the logarithmic barrier method \cite{nocedal}.
The logarithmic barrier is a simple and intuitive method  designed specifically to enforce inequality
constraints, which we leverage to enforce misclassification.
We compare the LogBarrier attack against current benchmark adversarial attacks
(using the Foolbox attack library \cite{foolbox}),
on several common datasets (MNIST \cite{mnist}, CIFAR10 \cite{cifar10},
ImageNet-1K \cite{imagenet}) and models. On average,
we show that the LogBarrier attack is comparable to current
state-of-the-art adversarial attacks. Moreover, we show that on challenging images (those that
require larger perturbations for misclassification), the LogBarrier attack
consistently outperforms other adversarial attacks. Indeed, we illustrate this
point by attacking models trained to be adversarially robust, and show that the
LogBarrier attack perturbs all images more efficiently than other attack
methods: the LogBarrier attack is able to perturb all images using a much smaller
perturbation size than that of other methods.

\section{Background material}
Adversarial examples arise in classification problems across multiple domains. The literature to date
has been concerned primarily with adversarial examples in image classification:
adversarial images appears no different (or only slightly
so) from an image correctly classified by a model, but despite this similarity,
are misclassified.

We let $\mathcal X$ be the space of images. Typically, pixel values are
scaled to be between 0 and 1, so that $\mathcal X$ is the unit box $[0,1]^M
\subset \R^M$. We let
$\mathcal Y$ be the space of labels. If the images can be one of $N$ classes,
$\mathcal Y$ is usually a subset of $\R^N$. Often $\mathcal Y$ is the
probability simplex, but not always. In this case, each element $y_i$ of a label $y$
correspond to the probability an image is of class $i$. Ground-truth labels are
then one-hot vectors.

A trained model, with fixed model weights $w$, is a map $f(\cdot;w):\mathcal X
\to \mathcal Y$. For brevity, in what follows we drop dependence on $w$.  For an input image $x$, the model's predicted
classification is the $\argmax$ of the model outputs. Given image-label pair
$(x,y)$, let $c$ be the index of the correct label (the $\argmax$ of $y$). A model is correct if
$\argmax f(x) = c$.

An adversarial image is a perturbation of the original
image, $x+\delta$,  such that the model misclassifies:
\begin{equation}
  \argmax f(x+\delta) \neq c
\end{equation}
The perturbation must be small in a certain sense: it must be small enough that
a human can still correctly classify the perturbed image. There are various
metrics for measuring the size of the perturbation. A common choice is the
$\ell_\infty$ norm (the max-norm); others use the (Euclidean) $\ell_2$ norm. If
perturbations must be sparse -- for example, if the attacker can only modify a
small portion of the total image -- then the count of non-zero elements in
$\delta$ may be used. Throughout this paper we let $m(\delta)$ be a generic metric on the
size of the perturbation $\delta$. Typically $m(\delta)$ is a specific $\ell_p$
norm, $m(\delta)=\Vert \delta \Vert_p$, such as the  Euclidean or  max norms.
Thus the problem of finding an adversarial image may be cast as an optimization
problem,
minimize the size of the perturbation subject to model misclassification.

The misclassification constraint is difficult to enforce, so a popular
alternative is to introduce a loss function $\mathcal L$. For example,
$\mathcal L$ could be the loss function used during model training. In this case
the loss measures the `correctness' of the model at an image $x$. If the loss is
large at a perturbed image $x+\delta$, then it is hoped that the image is also
misclassified. The loss function is then used as a proxy for
misclassification, which gives rise to the following indirect method for
finding adversarial examples:
\begin{equation}\label{eq:max-loss}
\begin{aligned}
& \underset{\delta}{\text{maximize}}
& & \mathcal L(x+\delta) \\
& \text{subject to}
& & m(\delta) \leq \varepsilon,
\end{aligned}
\end{equation}
maximize the loss subject to perturbations being smaller than a certain
threshold. 
The optimization approach taken by \eqref{eq:max-loss} is by far the most
popular method for finding adversarial examples. In one of the first papers on
this topic, Szegedy et al \cite{szegedy2013} proposed the Fast Signed
Gradient Method (FGSM), where $m(\delta)$ is the $\ell_\infty$ norm of the
perturbation $\delta$, and the
solution to \eqref{eq:max-loss} is approximated by taking one step in the signed
gradient direction. An iterative version with multiple steps, Iterative FGSM
(IFGSM) was
proposed in \cite{kurakin2016}, and remains the method of choice for adversarial
attacks measured in $\ell_\infty$. When perturbations are measured in $\ell_2$, 
\eqref{eq:max-loss} is solved with Projected Gradient Descent (PGD)
\cite{madry2017}. 

Fewer works have studied the adversarial optimization problem directly, i.e., without
a loss function.
In a seminal work, Carlini and Wagner \cite{carlini2017},
developed a \emph{targeted attack}, in which the adversarial distance is
minimized subject to a targeted misclassification. In a targeted
attack, not just any misclassification will do: the adversarial perturbation must
induce misclassification to a pre-specified target class. The Carlini-Wagner
attack (CW) incorporates the targeted misclassification constraint as a
penalty term into the objective function. The CW attack was able to overcome
many adversarial defence methods that had been thought to be effective, and
was an impetus for the adversarial research community's search for rigorous, theoretical
guarantees of adversarial robustness.

There is interest in gradient-free methods for finding adversarial examples. In this scenario, the attacker only has
access to the classification of the model, but not the model itself (nor the
model's gradients). In \cite{boundaryattack}, Brendal et al directly minimize
the $\ell_2$ adversarial distance while enforcing misclassification using a
gradient-free method. Their Boundary attack iteratively alternates between
minimizing
the perturbation size, and projecting the perturbation onto the classification
boundary. The projection step is approximated by locally sampling the model
decision near the classification boundary.

\section{The LogBarrier Attack}
\begin{figure}[t]
    \centering
    \includegraphics[width=0.5\textwidth]{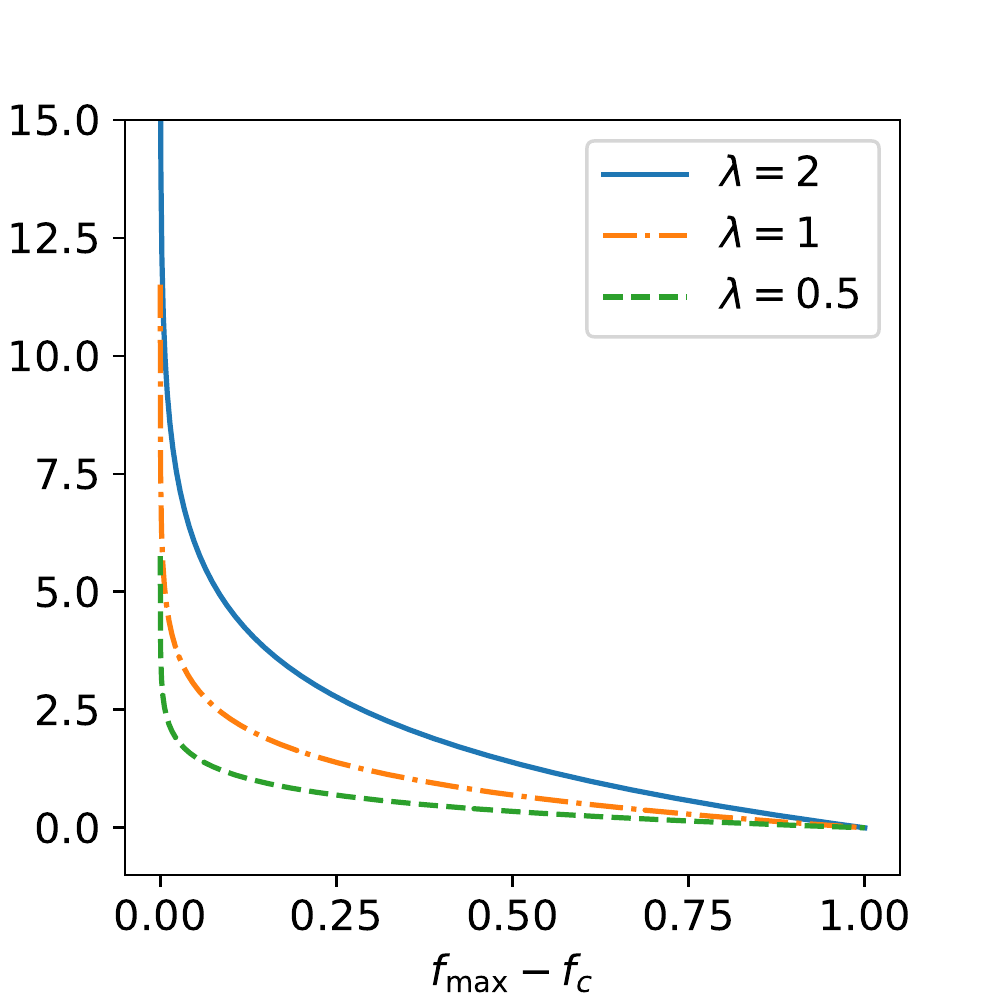}
    \caption{The logarithmic barrier function $\phi(\cdot):=- \lambda
    \log(\cdot)$ defined over $(0,1)$. As $\lambda$ decreases, the barrier
  becomes steeper, mimicking a hard  constraint.}
    \label{fig:logbar_motivation}
\end{figure}

We tackle the problem of finding (untargeted) adversarial examples by directly solving the
following optimization problem,
\begin{equation}\label{eq:best}
\begin{aligned}
& \underset{\delta}{\text{minimize}}
& & m(\delta) \\
& \text{subject to}
& & \argmax f(x+\delta) \neq c,
\end{aligned}
\end{equation}
that is, minimize the adversarial distance subject to misclassification. We use
the logarithmic barrier method
\cite{nocedal} to enforce  misclassification, as follows. We are given an image-label pair $(x,y)$ and correct label
$c=\argmax y$. Misclassification at an image $x$ occurs if there is at least one index of the
model's prediction with greater value than the prediction of the correct index:
\begin{equation}
  \max_{i\neq c} f_i(x) - f_c(x) > 0
\end{equation}
This is a necessary and sufficient condition for misclassification.
Thus, we rewrite \eqref{eq:best}:
\begin{equation}\label{eq:rewrite}
\begin{aligned}
& \underset{\delta}{\text{minimize}}
& & m(\delta) \\
& \text{subject to}
& & \max_{i\neq c} f_i(x+\delta) - f_c(x+\delta) >0.
\end{aligned}
\end{equation}
The barrier method is a standard tool in optimization for solving problems such
as \eqref{eq:rewrite} with inequality constraints. A complete discussion of the
method can be found in \cite{nocedal}. In the barrier method,
inequality constraints are incorporated into the objective function via a penalty
term, which is infinite if a constraint is violated. If a constraint is far from
being active, then the penalty term should be small. The negative logarithm is
an ideal choice:
\begin{equation}\label{eq:logbarrier}
  \min_\delta \, m(\delta) - \lambda \log\left(f_{\max}
- f_c \right)
\end{equation}
where we denote $f_{\max} := \max_i f_i(x+\delta)$ and $f_c:= f_c(x+\delta)$.
If the gap between $f_{\max}$ is much larger than $f_c$, the logarithmic barrier
term is small. However, as this gap shrinks, the penalty term
approaches infinity. Thus the penalty acts as a barrier, forcing an optimization
algorithm to search for solutions where the constraint is inactive. 
If \eqref{eq:logbarrier} is solved iteratively with smaller and smaller values of
$\lambda$, in the limit as $\lambda \to 0$, the solution to the original problem
\eqref{eq:rewrite} is recovered. (This argument can be made formal if desired,
using $\Gamma$-convergence \cite{braides2002gamma}.) See Figure \ref{fig:logbar_motivation}, where
the barrier function is plotted with decreasing values of $\lambda$. In the
limit as $\lambda\to 0$, the barrier becomes 0 if the constraint is satisfied,
and $\infty$ otherwise.

\subsection{Algorithm description}
We now give a precise description of our implementation of the log barrier
method for generating adversarial images.
The constraint $f_{\max} - f_c > 0$ can be viewed as a feasible set. Thus,
the algorithm begins by finding an initial feasible image: the original image
must be
perturbed so that it is
misclassified (not necessarily close to the original). There are several ways
to find a misclassified image. A simple approach would be to take another
natural image with a different label. However, we have found in practice that closer
initial images are generated by randomly perturbing the original image with increasing levels of noise
(\emph{e.g.} Standard Normal or Bernoulli) until it is misclassified. 
After each random perturbation, the image is projected back onto the set of
images in the $[0,1]^M$ box, via the projection $\mathcal P$.
This process is
briefly described in Algorithm \ref{alg:LogBarrierInit}.
Note that if the original image is already misclassified, no random
perturbation is performed, since the original image is already adversarial.

\begin{algorithm}[t]
   \caption{LogBarrier: Initialization}
   \label{alg:LogBarrierInit}
\begin{algorithmic}
   \STATE {\bfseries Input:} image $x \in \cX$, model $f(\cdot;w)$, $\rho$, step-size $h > 0$, $k_{\text{max}} \in \mathbb{N}$.
   \STATE {Initialize: }$B \sim \text{Bernoulli}(\rho) \in \cX$ or $B\sim
   \text{Normal}(0,1)$
   \FOR{$k=0$ {\bfseries to} $k_{\text{max}}$}
   \IF{$x$ misclassified}
   \STATE Exit \textbf{for-loop}
   \ELSE
   \STATE Sample $b$ from  $B$
   \STATE $x \leftarrow \mathcal{P}(x + h 1.01^k b)$
   \ENDIF
   \ENDFOR
\end{algorithmic}
\end{algorithm}

After an initial perturbation is found, we solve \eqref{eq:logbarrier} for a
fixed $\lambda$.
Various optimization methods algorithms may be used to solve \eqref{eq:logbarrier}. For
small- to medium-scale problems, variants of Newton's method are typically
preferred.
However, due to computational constraints, we chose to use gradient descent.
After each gradient descent step, we check to ensure that the updated
adversarial image remains in the $[0,1]^M$ box. If not, it is projected back
into the set of images with the projection $\mathcal P$.

It is possible that a gradient descent step moves the adversarial image so that
the image is correctly classified by the model. If this occurs, we simply
backtrack along the line between the current iterate and the previous iterate,
until we regain feasibility. To illustrate the backtracking procedure,
let $u^{(k)}$ be the previous iterate, and $\tilde{u}^{(k+1)}$ be a candidate
adversarial image which is now correctly classified. We continue backtracking the next iterate
via
\begin{equation}
  \tilde{u}^{(k+1)} \leftarrow \gamma \tilde{u}^{(k+1)} + (1-\gamma)u^{(k)},
\end{equation}
until the iterate is
misclassified. The hyper-parameter $\gamma \in (0,1)$ is a backtracking parameter. The accumulation point of the above
sequence is $u^{(k)}$. As a result, this process is guaranteed to terminate,
since the previous iterate is itself misclassified.
This backtracking procedure is sometimes necessary when iterates are very close
to the decision boundary. If the iterate is very close to the decision boundary,
then the gradient of the log barrier term is very large, and dominates the
update step.
Since the constraint set $f_{\max} - f_y > 0$ is not necessarily convex or even
fully connected, it is possible that the iterate could be sent far from the
previous iterate without maintaining misclassification. We rarely experience
this phenomenon in practice, but include the backtracking step as a safety.
An alternate approach (which we did not implement), more aligned with
traditional optimization techniques, would be instead to use a dynamic step size
rule such as the Armijo-Goldstein
condition \cite{armijo}.

\begin{figure}
    \centering
    \includegraphics[width=0.5\textwidth]{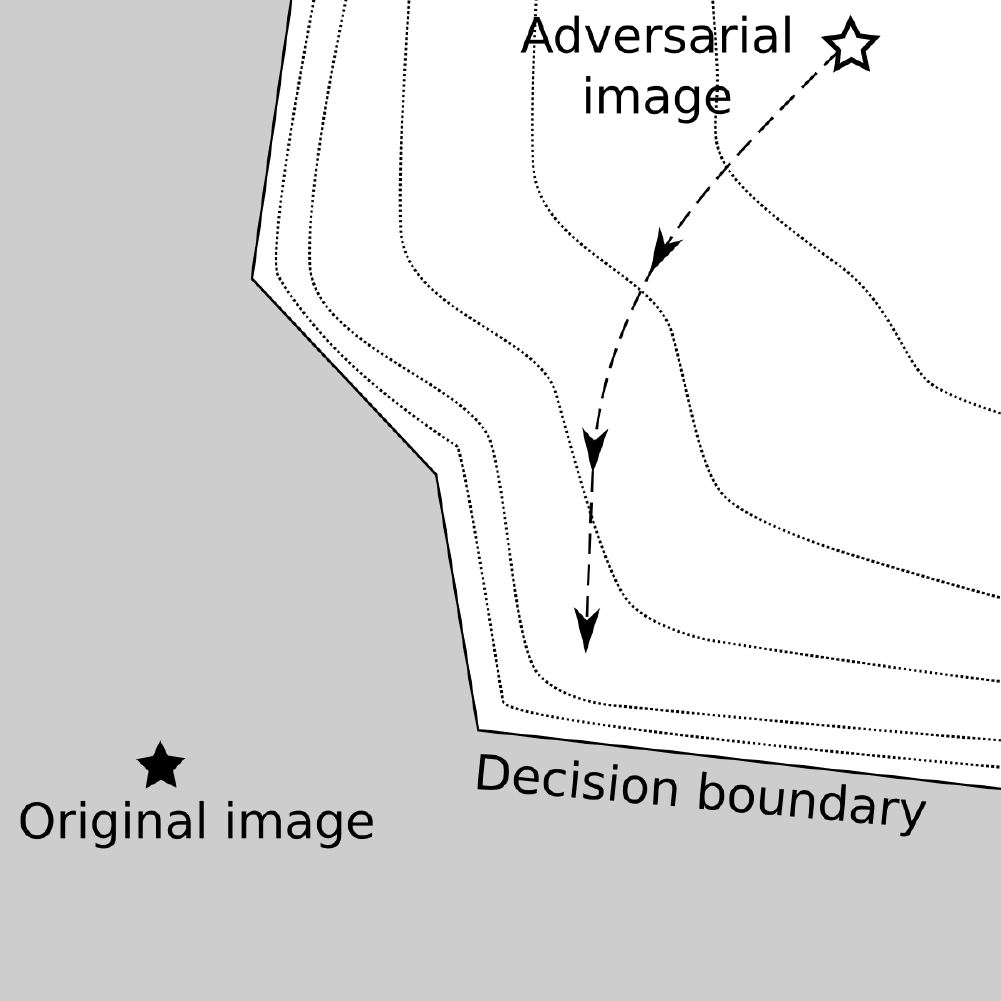}
    \caption{The central path taken by the LogBarrier
    attack. Dashed lines represent level sets of the logarithmic barrier
  function. As $\lambda$ decreases iterates approach the decision boundary.}\label{fig:diagram}
\end{figure}

\begin{algorithm}[t]
   \caption{LogBarrier attack}
   \label{alg:LogBarrierAttack}
\begin{algorithmic}
  \STATE {\bfseries Input:}  original image $x$, initial misclassified image $u^{(0)}$, model
  $f(\cdot;w)$, distance measure $m(\cdot)$ \\
   {\bfseries Hyperparameters:} backtrack factor $\gamma \in (0,1)$; initial penalty size
   $\lambda_0$; step size $h$; $\lambda$ shrink factor $\beta \in (0,1)$;
   termination threshold $\varepsilon > 0$; and maximum iterations  $K_{\text{outer}}, J_{\text{inner}} \in \mathbb{N}$.
   \FOR{$k=0$ {\bfseries to} $K_{\text{outer}}$}
   \STATE $\lambda_k = \lambda_0\beta^k$
   \FOR{$j=0$ {\bfseries to} $J_{\text{inner}}$}
   \STATE \begin{align*}
     & \tilde{u}^{(j+1)} \leftarrow u^{(j)} - h\nabla \left(m(u^{(j)}-x) + \lambda_k \phi(u^{(j)}) \right) \\
       & u^{(j+1)} \leftarrow \mathcal{P}\left(\tilde{u}^{(j+1)}\right)
       \quad\text{project onto $[0,1]^M$}
   \end{align*}
   \WHILE{$u^{(j+1)}$ {\bfseries not} misclassified}
   \STATE $$ u^{(j+1)} \leftarrow \gamma u^{(j+1)} + (1-\gamma)u^{(j)}   $$
   \ENDWHILE
   \IF{$\|u^{(j+1)} - u^{(j)} \| \leq \varepsilon$}
   \STATE break
   \ENDIF
   \ENDFOR
   \ENDFOR
\end{algorithmic}
\end{algorithm}

The gradient descent algorithm comprises a series of iterates in an inner loop. 
Recall that as $\lambda\to 0$, the log barrier problem approaches the original
problem \eqref{eq:rewrite}.
Thus, we shrink $\lambda$ by a certain factor and repeat the procedure
again, iterating in a series of outer loops (of course, initializing now with
previous iterate). As $\lambda$ shrinks, the solutions
to \eqref{eq:logbarrier} approach the decision boundary.
In each inner loop, if the iterates fail to move less than some threshold value
$\varepsilon$, we move onto the next outer loop.
The path taken by the iterates of the outer loop is called the \emph{central
path}, illustrated in Figure  \ref{fig:diagram}.  

The LogBarrier attack pseudocode is presented in Algorithm
\ref{alg:LogBarrierAttack}. For brevity we write the log barrier $\phi(u):= -\log(
\max_i f_i(u) - f_c(u))$.
We remark that the LogBarrier attack can be improved by running the
method several times, with different random initializations (although we do not
implement this here).

%\subsection{Motivation and potential drawbacks}
%The logarithmic barrier method is a conventional tool in optimization and can be applied to convex and non-convex problems alike; a complete discussion can be found in [cite Boyd]. The intuition of the behaviour is displayed in the following graph, where changing the strength of the ``$a$" parameter dictates the strength of the barrier. 
%
%

The literature on adversarial perturbations primarily focuses on perturbations
measured in the $\ell_2$ and $\ell_\infty$ norms.  For perturbations measured in
the $\ell_2$ norm, we set the distance measure to be the squared Euclidean norm,
$m(\delta) = \Vert \delta \Vert^2_2$. When perturbations are measured in the
$\ell_\infty$ norm, we do not use the max-norm directly as a measure, due to the
fact that the $\ell_\infty$ norm is non-smooth with sparse subgradients. Instead, we
use the following approximation  of the $\ell_\infty$ norm
\cite{lange2014applications},
\begin{align*}
    \|\delta\|_\infty &= \max_{i=1,\ldots,N} |\delta_i| \\
    &\approx \dfrac{\sum_{i=1}^N |\delta_i| \exp(\alpha|\delta_i|)
    }{\sum_{i=1}^N \exp(\alpha|\delta_i|) },
\end{align*}
where $\alpha > 0$. As $\alpha \to \infty$ the $\ell_\infty$ norm is recovered. 

\subsubsection*{Algorithm hyper-parameters}

Like many optimization routines, the logarithmic barrier method has several
hyper-parameters. However, because our implementation is parallelized, we have
found that the tuning process is relatively quick. For the $\ell_\infty$ attack,
our default parameters are $\varepsilon = 10^{-6}, h = 0.1, \beta = 0.75, \gamma
= 0.5, \lambda_0 = 0.1, K_{\text{outer}} = 25$, and  $J_{\text{inner}} = 1000$.
For $\ell_2$, we set $h = 5 \cdot 10^{-3}$ with $K_{\text{outer}}=15$ and
$J_{\text{inner}}=200$; the rest are the same as in the $\ell_\infty$ case.

For the initialization procedure, we have $k_{\max} = 10^3$ and $h = 5\cdot
10^{-4}$. If attacking in $\ell_2$, we initialize using the Standard Normal
distribution. Else, for $\ell_\infty$, we use the Bernoulli initialization with
$\rho = 0.01$. 

\subsubsection*{Top5 misclassification}
The LogBarrier attack may be generalized to enforce Top5 misclassification as
well. In this case, the misclassification constraint is that $f_{(k)}(x+\delta) -
f_c(x+\delta)>0, \,k=1,\dots, 5$, where now $(k)$ is the index of sorted model
outputs. (In other words, $f_{(1)}=\max_i f_i$, and $f_{(2)}$ is the second-largest model
output, and so forth.) We then set the barrier function to be $-\sum_{k=1}^5
\log(f_{(k)}-f_c)$. In this scenario, the LogBarrier attack is initialized with an
image that is not classified in the Top5.
 
\section{Experimental results}
We compare the LogBarrier attack with current state-of-the-art adversarial
attacks on three benchmark datasets: MNIST \cite{mnist},
CIFAR10 \cite{cifar10}, and ImageNet-1K \cite{imagenet}. On MNIST and CIFAR10, we attack 1000 randomly chosen
images; on ImageNet-1K we attack 500 randomly selected images, due to
computational constraints.  On ImageNet-1K, we use the Top5 version of the
LogBarrier attack.

All other attack methods are implemented using the adversarial attack library
Foolbox \cite{foolbox}. For adversarial attacks measured in $\ell_2$, we compare the LogBarrier attack against
Projected Gradient Descent (PGD) \cite{madry2017}, the Carlini-Wagner attack
(CW) \cite{carlini2017}, and the Boundary attack (BA) \cite{boundaryattack}.
These three attacks all very strong, and consistently perform well in adversarial
attack competitions.
When measured in $\ell_\infty$, we compare against IFGSM \cite{kurakin2016}, the
current state-of-the-art.
We leave Foolbox hyper-parameters to their defaults,
except the number of iterations in the Boundary attack, which we set to a maximum of 5000 iterations.
 
\subsection{Undefended networks}\label{sec:undefended}

\begin{table}%[H]
\caption{Percent misclassification of the networks at a specified perturbation
size, for attacks measured in $\ell_2$.  Because we are measuring the strength
of adversarial attacks, at a given adversarial distance, a higher percentage
misclassified is better. }
\begin{adjustbox}{width=.6\columnwidth}
\centering
\begin{tabular}{l||c|c|c|c}
  & \multirow{2}{*}{MNIST}    & \multicolumn{2}{c|}{CIFAR10} &
  \multirow{2}{*}{Imagenet-1K} \\
& & AllCNN & ResNeXt34 &  \\  

$\|\delta\|_2$ & 2.3 & $120/255$ &
$120/255$ &
$1$  \\ \hline\hline
LogBarrier & \textbf{99.10}  & \textbf{98.70} & \textbf{99.90} & \textbf{98.40}  \\
CW & 98.50 & 97.30 & 90.40 & 74.86 \\
PGD & 52.58  & 86.60 & 59.80 & 90.00 \\
BA & 97.20 & \textbf{98.70} & 99.60 & 48.80
\end{tabular}
\end{adjustbox}
\label{tab: l2-undef}
\end{table}

\begin{table}%[H]
\caption{Percent misclassification of the networks at a specified perturbation
size, for attacks measured in $\ell_\infty$.  Higher percentage misclassified is better. }
\begin{adjustbox}{width=.6\columnwidth}
\centering
\begin{tabular}{l||c|c|c|c}
  & \multirow{2}{*}{MNIST}    & \multicolumn{2}{c|}{CIFAR10} &
  \multirow{2}{*}{Imagenet-1K} \\
& & AllCNN & ResNeXt34 &  \\  
 $\|\delta\|_\infty$ & 0.3 & $8/255$ &
$8 / 255$ &
$8 / 255$ \\ \hline \hline
LogBarrier & \textbf{94.80}  & \textbf{100} & \textbf{98.70}   & 95.20  \\
IFGSM & 73.40 & 93.1 & 75.80 & \textbf{99.60 } \\
\end{tabular}
\end{adjustbox}
\label{tab: linf-undef}
\end{table}

\begin{table*}[!t]
\caption{Adversarial attacks perturbation statistics in the $\ell_2$ norm. We
  report the mean and variance of the adversarial distance on a subsample of the test dataset.
Lower values are better.}
\centering
\begin{tabular}{l||c c |c c |c c |c c }
  \multirow{3}{*}{} & \multicolumn{2}{c|}{\multirow{2}{*}{MNIST}}  &
  \multicolumn{4}{c|}{CIFAR10} &
  \multicolumn{2}{c}{\multirow{2}{*}{ImageNet-1K}} \\
  & &  & \multicolumn{2}{c}{AllCNN} & \multicolumn{2}{c|}{ResNeXt34} &  &  \\ 
  & $\mu$ & $\sigma^2$ & $\mu$ & $\sigma^2$ & $\mu$ & $\sigma^2$ & $\mu$ &
  $\sigma^2$ \\ \hline \hline
LogBarrier &  1.29 & \num{1.98e-1}  & \textbf{\num{1.63e-1}} &
\textbf{\num{1.12e-2}}  & \num{1.21e-1} & \textbf{\num{6.68e-3}} &
\textbf{\num{3.82e-1}} & \textbf{\num{6.87E-2}} \\
CW & \textbf{1.27} & \textbf{\num{1.96E-1}}  & \num{1.72E-1} & \num{8.57E-2}
& \num{2.39E-1} & \num{1.87E-1} & \num{8.86E-1} &  1.59 \\
PGD & 2.54 & 2.53  & \num{3.18E-1} & \num{3.49E-1}    & \num{6.88E-1} & 1.15&
\num{4.21E-1} &  \num{3.16E-1} \\
BA & 1.41 & \num{2.11E-1}  & \textbf{\num{1.63E-1}} &  \num{1.36E-2}  & \textbf{\num{1.11E-1}} &
\num{7.396E-3} &  1.55 & 3.31 \\
\end{tabular}
\label{tab: meanvar-l2}
\end{table*}

\begin{table*}[!t]
\begin{center}
\caption{Adversarial attacks perturbation statistics in the $\ell_\infty$ norm.
We report the mean and variance of the adversarial attack distance for each
method on a subsample of the test dataset. Lower values are better.}
\label{tab: meanvar-linf}
\begin{tabular}{l||c c |c c |c c |c c }
  \multirow{3}{*}{} & \multicolumn{2}{c|}{\multirow{2}{*}{MNIST}}  &
  \multicolumn{4}{c|}{CIFAR10} &
  \multicolumn{2}{c}{\multirow{2}{*}{ImageNet-1K}} \\
  & &  & \multicolumn{2}{c}{AllCNN} & \multicolumn{2}{c|}{ResNeXt34} &  &  \\ 
  & $\mu$ & $\sigma^2$ & $\mu$ & $\sigma^2$ & $\mu$ & $\sigma^2$ & $\mu$ &
  $\sigma^2$ \\ \hline \hline
  LogBarrier &  \textbf{\num{1.57E-1}} &  \textbf{\num{7.43E-3}} &
  \textbf{\num{6.16E-3}} & \textbf{\num{1.3E-5}} & \textbf{\num{5.14E-3}} &
  \textbf{\num{3.20E-5}} & \num{1.27E-2} & \num{1.46E-3} \\
  IFGSM & \num{2.49E-1} & \num{3.4E-2}  & \num{1.14E-2} & \num{6.93E-4}   &
  \num{2.70E-2} & \num{2.07E-3} & \textbf{\num{2.38E-3}} & \textbf{\num{1.30E-5}} \\
\end{tabular}
\end{center}
\end{table*}
We first study the LogBarrier attack on networks that have not been trained to be
adversarially robust. For MNIST, we use the network described in
\cite{carlini2017,papernot2016}. On CIFAR10,
we consider two networks: AllCNN \cite{allcnn}, a shallow convolutional
network; and a
ResNeXt34 (2x32) \cite{resnext}, a much deeper network residual network.
Finally, for ImageNet-1K, we use a pre-trained ResNet50 \cite{resnet} available
for download on the PyTorch website. %\\ \hfill 

Tables \ref{tab: l2-undef} and \ref{tab: linf-undef} report the percentage
misclassified, for each attack at a fixed perturbation size. A strong attack should
have a high misclassification rate. In the tables, the perturbation
size is chosen to agree with attack thresholds commonly reported in the
adversarial literature. Measured in Euclidean
norm, we see that the LogBarrier attack is the strongest on all datasets
and models. Measured in the max-norm, the LogBarrier outperforms IFGSM on all
datasets and models, except on ImageNet-1K where the
difference is slight.

We also report the mean and variance of the adversarial attack distances,
measured in $\ell_2$ and $\ell_\infty$, in Tables
\ref{tab: meanvar-l2} and \ref{tab: meanvar-linf} respectively. A strong adversarial attack
should have a small mean adversarial distance, and a small variance. Small
variance is necessary to ensure precision of the attack method. A strong attack
method should be able to consistently find close adversarial examples. Table
\ref{tab: meanvar-l2} demonstrates that, measured in $\ell_2$, the LogBarrier attack is either the
first ranked attack, or a close second. When measured in $\ell_\infty$, the
LogBarrier attack significantly outperforms IFGSM on all datasets and models, except
ImageNet-1K.

\begin{figure*}
    \centering
    \begin{subfigure}[b]{0.45\textwidth}
        \includegraphics[width=\textwidth]{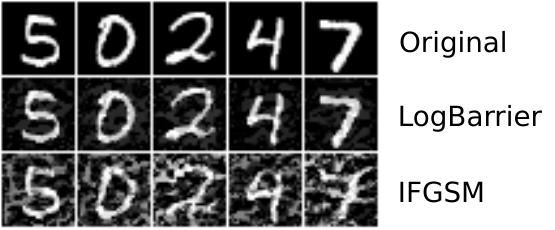}
        \caption{$\ell_\infty$ attacks on MNIST}
        \label{fig:mnist_pics}
    \end{subfigure}
    \hspace{2em} %add desired spacing between images, e. g. ~, \quad, \qquad, \hfill etc. 
    \begin{subfigure}[b]{0.45\textwidth}
        \includegraphics[width=\textwidth]{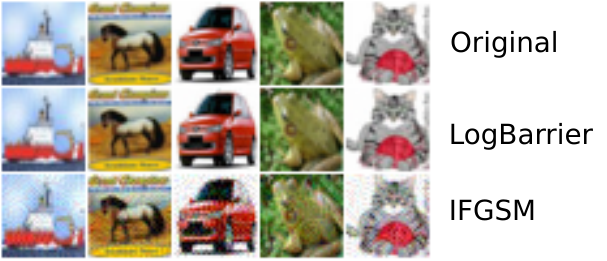}
        \caption{$\ell_\infty$ attacks on CIFAR10}
        \label{fig:cifar10_pics}
    \end{subfigure}
    \caption{Adversarial images for  $\ell_\infty$ perturbations, generated by
      the LogBarrier and IFGSM adversarial attacks, compared against the
      original clean image. Where IFGSM has
    difficulties finding adversarial images, the LogBarrier method succeeds:
    LogBarrier adversarial images are visibly less distorted than IFGSM
    adversarial images.
  }\label{fig:pics}
\end{figure*}

For illustration, we show examples of adversarial images from the IFGSM and
LogBarrier attacks in Figure \ref{fig:pics}. On images where IFGSM requires a
large distance to adversarially perturb, the LogBarrier attack produces visibly
less distorted images.

\begin{figure*}
    \centering
    \begin{subfigure}[b]{0.4\textwidth}
        \includegraphics[width=\textwidth]{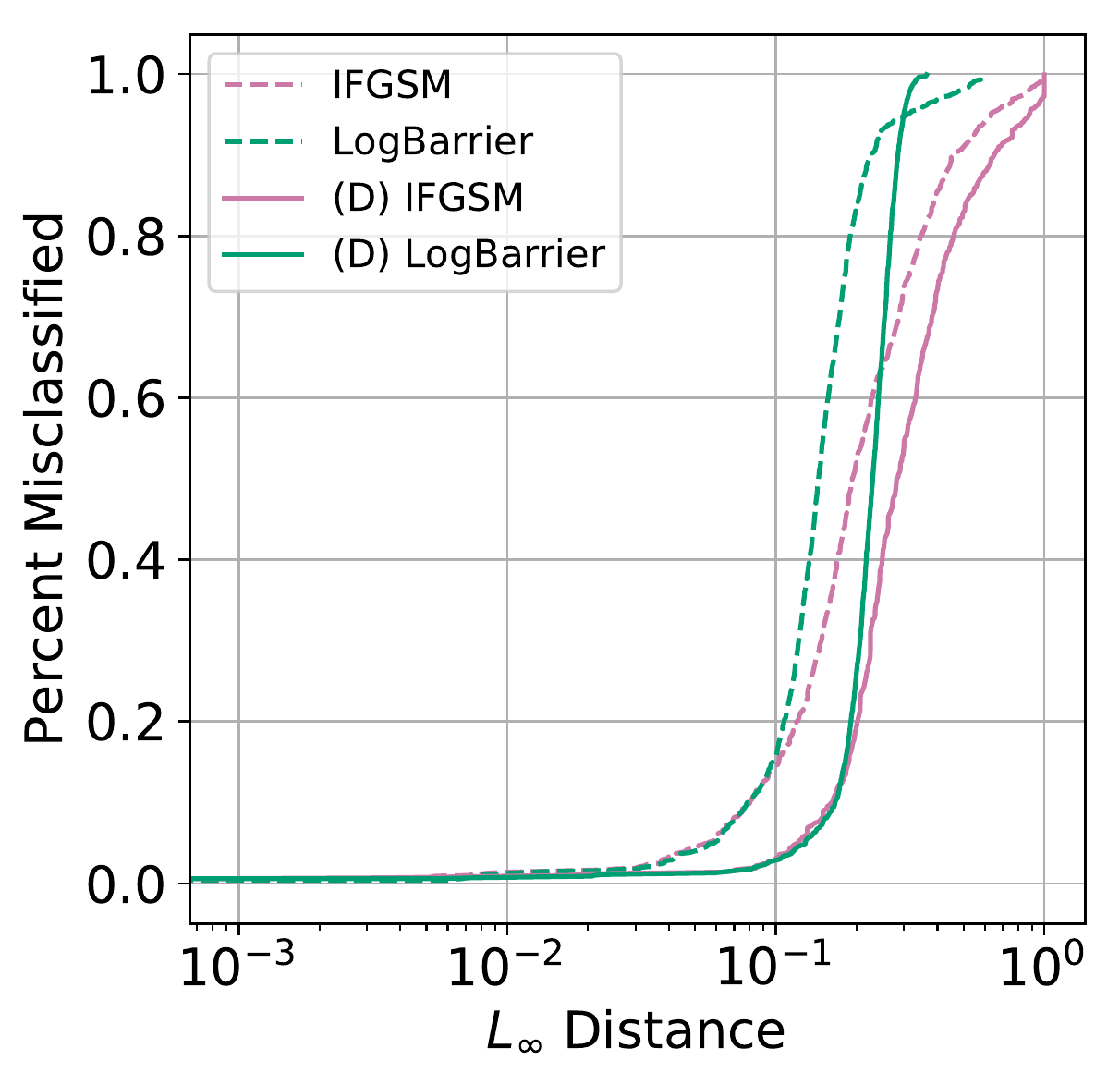}
        \caption{MNIST}
        \label{fig:mnist_overlay}
    \end{subfigure}
    \hspace{2em} %add desired spacing between images, e. g. ~, \quad, \qquad, \hfill etc. 
    \begin{subfigure}[b]{0.4\textwidth}
        \includegraphics[width=\textwidth]{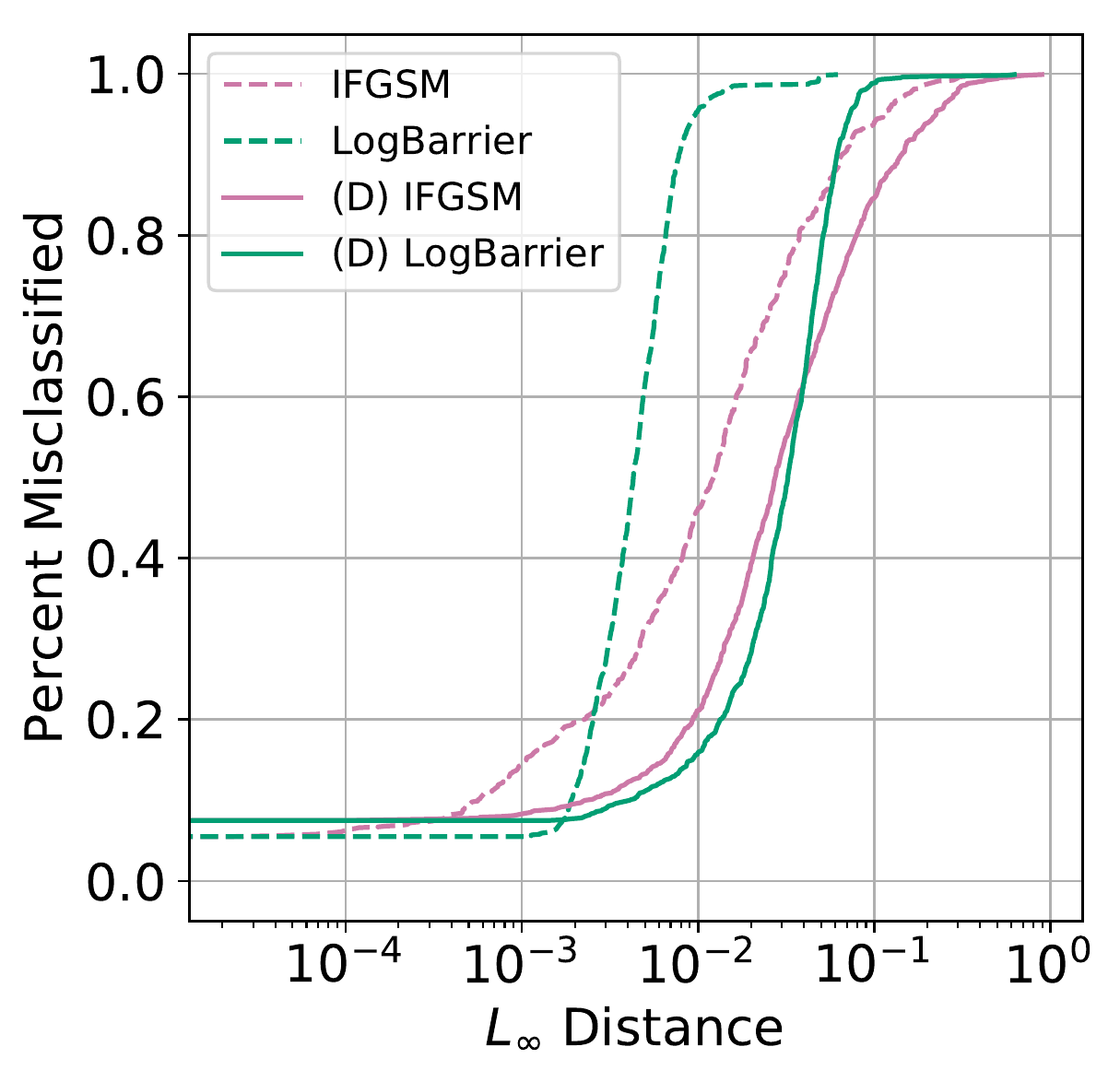}
        \caption{CIFAR10}
        \label{fig:cifar10_overlay}
    \end{subfigure}
    \caption{Overlay of attack curves, measured in $\ell_\infty$, on (a) MNIST and (b)  CIFAR10 networks.
      Two types of networks are compared: an undefended network, and 
      a defended network (denoted (D)), trained using the same architecture as the
      undefended network with adversarial training. The LogBarrier
      attack requires a smaller adversarial distance to attack all images,
      compared to IFGSM.}\label{fig:curves}
\end{figure*}

\subsection{Defended networks}

In this section we turn to attacking adversarially defended networks.
We first consider two defence strategies: gradient obfuscation
\cite{athalye2018}, and
multi-step adversarial training as described in Madry et al \cite{madry2017}. We study
these two strategies on the MNIST and ResNeXt34 networks used in Section
\ref{sec:undefended}. We limit ourselves to studying defence methods for attacks
in the $\ell_\infty$ norm. Attacks are performed on the same 1000 randomly
selected images as the previous section. 
Finally, we also
test our attack on a MNIST model trained with Convex Adversarial Polytope \cite{convexpolytope}
training, the current state-of-the-art defence method on MNIST.

\subsubsection*{Gradient Obfuscation}
Although discredited as a defence method \cite{athalye2018}, gradient
obfuscation is a hurdle any newly proposed adversarial attack method must be
able to surmount. We implement gradient obfuscation by increasing the
temperature on the softmax function computing model probabilities from model
logits. As the softmax temperature increases, the size of the gradients of the model
probabilities approaches zero, because the model probabilities approach one-hot
vectors. Although the decision boundary of the model does
not change, many adversarial attack algorithms have difficulty generating
adversarial examples when model gradients are small.

In Tables \ref{tab: def-mnist} and \ref{tab: def-cifar10} we show that the
LogBarrier attack easily overcomes gradient obfuscation, on both CIFAR10 and
MNIST models.
The reason that the LogBarrier method is able to overcome gradient obfuscation is
simple: away from the decision boundary, the logarithmic barrier term is not
active (indeed, it is nearly zero). Thus the LogBarrier algorithm focuses on
minimizing the adversarial distance, until it is very close to the decision
boundary, at which point the barrier term activates.
In contrast, because IFGSM is a local method, if model gradients are small, it has a
difficult time climbing the loss landscape, and is not able to generate
adversarial images.

\begin{table*}[!t]
%\begin{center}
\caption{Defence strategies on MNIST. We report the percentage misclassified at
$\ell_\infty$ adversarial magnitudes $\Vert \delta \Vert_\infty=0.1$ and  0.3;
higher is better. We also report the attack magnitude needed to perturb 90\% of
the images (the 90\% quantile of attacks, written $q(90\%)$).  'NA'
indicates that the attack failed.}
\begin{adjustbox}{width=\textwidth}
\centering
\begin{tabular}{l||c c c|c c c|c c c}
  & \multicolumn{3}{c|}{Undefended} & \multicolumn{3}{c|}{Obfuscated $(T = 2)$}
  &\multicolumn{3}{c}{Adversarial training} \\ 
  & $\Vert \delta \Vert_\infty = 0.1$ & $\Vert \delta \Vert_\infty = 0.3$ &
  $q(90\%)$ &
$\Vert \delta \Vert_\infty = 0.1$ & $\Vert \delta \Vert_\infty = 0.3$ &
  $q(90\%)$ &
$\Vert \delta \Vert_\infty = 0.1$ & $\Vert \delta \Vert_\infty = 0.3$ &
  $q(90\%)$ \\
  \hline
  \hline
  LogBarrier & \textbf{15.70}& \textbf{94.80}& \textbf{\num{2.27E-1}} &
  \textbf{18.30}& \textbf{99.80}& \textbf{\num{1.95E-1}} & 2.90 & \textbf{95.40} &
\textbf{\num{2.85E-1}}  \\
IFGSM & 12.40 & 62.5 & \num{4.60E-1} & 8.60 &  32.90 & NA & \textbf{3.00} &
53.80 & \num{6.51E-1}  \\
\end{tabular}
%\end{center}
\end{adjustbox}
\label{tab: def-mnist}
\end{table*}

\begin{table*}[!t]
\caption{Defence strategies on ResNeXt34 on the  CIFAR10 dataset. We report the
percentage misclassified at $\ell_\infty$ adversarial magnitudes of $\Vert \delta
\Vert_\infty=4/255$ and
$8/255$, and the magnitude required to perturb 90\% of test images. If the
adversarial attack was unsuccessful, we report NA.}
\begin{adjustbox}{width=\textwidth}
\centering
\begin{tabular}{l||c c c|c c c|c c c}
  & \multicolumn{3}{c|}{Undefended} & \multicolumn{3}{c|}{Obfuscated $(T = 20)$}
  &\multicolumn{3}{c}{Adversarial training} \\ 
  & $\Vert \delta \Vert_\infty = \frac{4}{255}$ & $\Vert \delta \Vert_\infty =
  \frac{8}{255}$ &
  $q(90\%)$ &
  $\Vert \delta \Vert_\infty =\frac{4}{255}$ & $\Vert \delta \Vert_\infty =
  \frac{8}{255}$ &
  $q(90\%)$ &
  $\Vert \delta \Vert_\infty = \frac{4}{255}$ & $\Vert \delta \Vert_\infty =
  \frac{8}{255}$ &
  $q(90\%)$ \B \\
  \hline
  \hline
  LogBarrier & \textbf{98.40}& \textbf{98.70}& \textbf{\num{7.79E-3}} & \textbf{47.60}
  &\textbf{54.40} &\textbf{\num{1.53E-1}} & 23.40 & 48.10 & \textbf{\num{9.58E-2}}   \\
IFGSM & 58.30 & 75.80 & \num{6.56E-2} & 36.90 & 43.90 & NA  & \textbf{31.60} &
\textbf{54.90} & \num{1.38E-1} \\
\end{tabular}
\end{adjustbox}
\label{tab: def-cifar10}
\end{table*}

\subsubsection*{Adversarial Training}

Adversarial training is a popular method for defending against adversarial
attacks.
We test the LogBarrier attack on networks trained with multi-step adversarial
training in the $\ell_\infty$ norm, as presented in Madry et al \cite{madry2017}. 
Our results are shown in Tables \ref{tab: def-mnist} and \ref{tab:
def-cifar10}. We also plot defence curves of the LogBarrier and IFGSM attacks on
defended and undefended models in Figures \ref{fig:mnist_overlay} and
\ref{fig:cifar10_overlay}, for respectively MNIST and CIFAR10.

On MNIST, we did 
not observe a reduction in test accuracy on clean images with adversarially
trained models compared to undefended models.  As expected, adversarial training hinders both LogBarrier and IFGSM from
finding adversarial images at very small distances. However, we see that the
LogBarrier attack is able to attack all images with nearly the same distance in
both the defended and undefended models. In contrast, IFGSM requires a very
large adversarial distance to attack all images on the defended model, as shown
in Figure \ref{fig:mnist_overlay}. 
That is, adversarial training \emph{does not significantly reduce the empirical 
distance required to perturb all images}, when the LogBarrier attack is used. The point is
illustrated in Table \ref{tab: def-mnist}, where we report the distance required
to perturb 90\% of all images. The LogBarrier attack requires an adversarial
distance of 0.22 on the undefended MNIST model, and 0.29 on the defended MNIST
model, to perturb 90\% of all images. In contrast, IFGSM requires a distance of
0.46 on the undefended model, but 0.65 on the defended model.

On CIFAR10, we observe the same behaviour, although the phenomenon is less
pronounced. As shown in Table \ref{tab: def-cifar10} and Figure
\ref{fig:cifar10_overlay}, the LogBarrier attack requires a smaller adversarial
distance to perturb all images than IFGSM. Notably, the LogBarrier attack on the
defended network is able to attack all images with a smaller adversarial
distance than even IFGSM on the undefended network.

\subsubsection*{Against the Convex Adversarial Polytope}
Finally, we use the LogBarrier attack on a provable defence strategy, the Convex
Adversarial Polytope \cite{convexpolytope}. The Convex Adversarial Polytope is a
method for training a model to guarantee that no more than a certain percentage of
images may be attacked at a given adversarial distance. We chose to attack
the defended MNIST network in \cite{convexpolytope}, which is guaranteed to have
no more than 5.82\% misclassification at perturbation size $\Vert \delta
\Vert_\infty=0.1$. We validated this theoretical guarantee with both the
LogBarrier attack and IFGSM, and found that both methods were unable to perturb
more than 3\% of test images at distance 0.1.

\section{Discussion}
We have presented a new adversarial attack that uses a traditional method from the
optimization literature, namely the logarithmic barrier method. The LogBarrier
attack is effective in both the $\ell_\infty$ and $\ell_2$ norms. The LogBarrier
attack \emph{directly} solves the optimization problem
posed by the very definition of adversarial images; i.e., find an image close
to an original image,  while being misclassified by a network. This is in contrast to
many other adversarial attack problems (such as PGD or IFGSM), 
which attempt to maximize a loss function as a proxy to the true adversarial
optimization problem. Whereas loss-based adversarial attacks start locally at or near the
original image, the LogBarrier attack begins far from the original image.
In this sense, the LogBarrier attack is similar in spirit to the Boundary attack \cite{boundaryattack}: both the LogBarrier
attack and the Boundary attack begin with a misclassified image, and iteratively
move the image
closer to the original image, while maintaining misclassification. The
LogBarrier attack is a gradient-based attack: to enforce misclassification,
gradients of the logarithmic barrier are required. In contrast, the Boundary
attack is gradient-free, and uses rejection sampling to enforce
misclassification. Although the LogBarrier attack uses gradients, we have shown
that it is not impeded by gradient obfuscation, a common drawback to other
gradient-based attacks. Because the LogBarrier attack is able to use gradients,
it is typically  faster than the Boundary attack.

The LogBarrier attack may be used as an effective tool to validate claims of
adversarial robustness. We have shown that one strength of the LogBarrier attack
is its ability to attack all images in a test set, using a fairly small maximum
adversarial distance compared to other attacks. In other words, the LogBarrier
attack estimates the mean adversarial distance with high precision. Using the
LogBarrier attack, we have raised questions about the robustness of multi-step
adversarial training \cite{madry2017}. For instance, on MNIST, we showed that
multi-step adversarial training did not significantly improve the necessary
distance required to perturb all test images, relative to an undefended model.
For adversarially trained models on CIFAR10, we showed that the necessary
distance to perturb all images is significantly smaller than
the estimate provided by IFGSM.  This is further motivation for the development
of rigorous, theoretical guarantees of model robustness.

%{\small
%\bibliographystyle{ieee}
%\bibliography{refs}
%}
%\bibliographystyle{alpha}
%\bibliography{refs}
\newcommand{\etalchar}[1]{$^{#1}$}

\end{document}